\pdfoutput=1

\documentclass[11pt]{article}

\usepackage{acl}
\usepackage{graphicx}
\usepackage{times}
\usepackage{latexsym}
\usepackage{tablefootnote}
\usepackage{longtable}
\usepackage{arydshln}
\usepackage{times}
\usepackage{latexsym}
\usepackage{tablefootnote}
\usepackage{longtable}
\usepackage{arydshln}

\usepackage[T1]{fontenc}

\usepackage[utf8]{inputenc}

\usepackage{microtype}

\title{High Recall Data-to-text Generation with Progressive Edit}

\author{ChoongHan Kim\\
  Graduate School of Artificial Intelligence\\ POSTECH, Pohang, South Korea
  \\
   \\
  \texttt{choonghankim@postech.ac.kr} \\\And
 Gary Geunbae Lee \\
  Graduate School of Artificial Intelligence
  \\ POSTECH, Pohang, South Korea
   \\
   \\
  \texttt{gblee@postech.ac.kr} \\}

\begin{document}
\maketitle
\begin{abstract}

Data-to-text (D2T) generation is the task of generating texts from structured inputs. We observed that when the same target sentence was repeated twice, Transformer (T5) based model generates an output made up of asymmetric sentences from structured inputs. In other words, these sentences were different in length and quality. We call this phenomenon "Asymmetric Generation" and we exploit this in D2T generation. Once asymmetric sentences are generated, we add the first part of the output with a no-repeated-target. As this goes through progressive edit (ProEdit), the recall increases. Hence, this method better covers structured inputs than before editing. ProEdit is a simple but effective way to improve performance in D2T generation and it achieves the new state-of-the-art result on the ToTTo dataset. 
\end{abstract}
 
\section{Introduction}

Data-to-text (D2T) generation is the task of generating texts from structured inputs \citep{reiter1997building}. Previous attempts to solve this task can be classified according to whether separate stages are adopted or not. For example, one method is to generate structured inputs in correct order first, then realize the whole sentence \citep{puduppully2019data,wang2021sketch,su2021plan}. Another is to generate the whole sentence in an end-to-end (E2E) manner using Copy mechanism \citep{gu2016incorporating,see2017get} or just pre-trained models \citep{kale2020text}.

The two methods each have their advantages and disadvantages. Methods that have separate stages generate more confident texts than E2E models since separate stages first produce entities from structured inputs. However, the output sentence can be awkward or the overall performance may be worse than that of E2E models. This occurs because the generated output of the first stage differs from the gold label that the second stage expects. If the first stage produces a slightly incorrect result, the second stage takes over and increases the error of subsequent sentences. This is often referred to as error propagation. E2E models are free from this vulnerability but it could omit the important entities from structured inputs. To include these important entities, a separate module may be added to E2E models (e.g. Copy mechanism module), but this could generate awkward sentences and degrade system integrity. For this reason, we consider the usage of a pre-trained model without additional modules. A pre-trained E2E Transformer model (e.g. T5 \citep{raffel2019exploring}) shows a competitive performance for D2T tasks \citep{kale2020text}.

\begin{figure}[t]
\begin{center}
\includegraphics[width=1\linewidth]{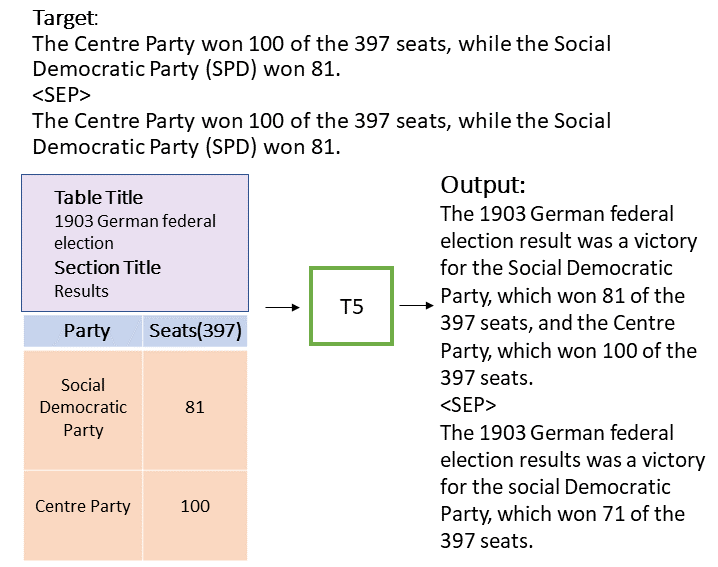}
\end{center}
  \caption{An example of generating asymmetric sentences. }
\label{fig:asymmetric}
\end{figure}

Omitting important entities from structured inputs is related to recall values. In D2T, the recall value is a metric that considers not only the target sentence, but also the structured inputs. A high recall indicates that more structured inputs are included. This metric is described in detail in Section 2.


\begin{figure*}[htbp]
\begin{center}
\includegraphics[width=1\linewidth]{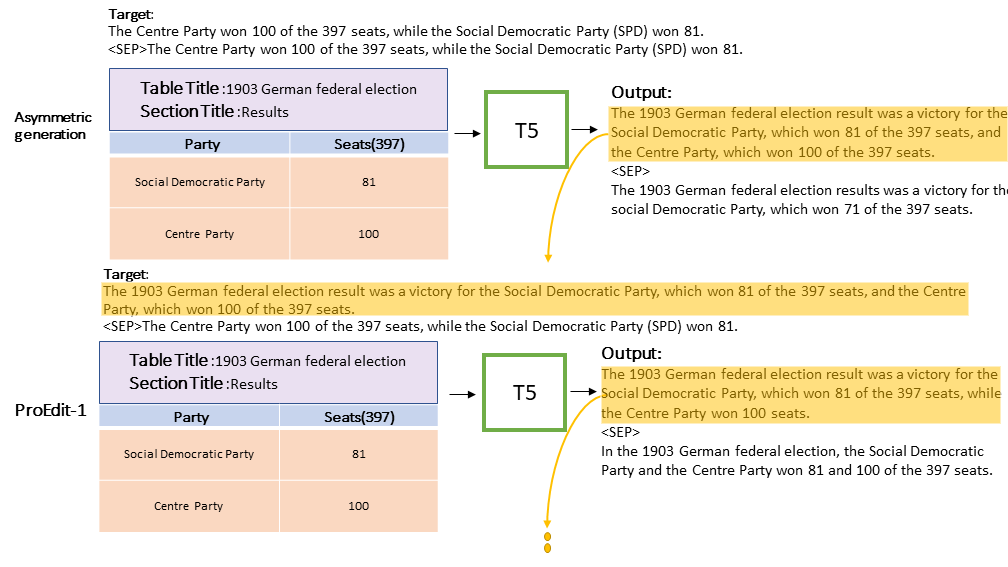}
\end{center}
  \caption{Progressive edit (ProEdit) of the output sentence. Only first parts of the output of asymmetric generation is used for next stage.}
\label{fig:proedit-3}
\end{figure*}

When the same target sentence is repeated, but divided by a special token (i.e. "\textit{target\_1} <SEP> \textit{target\_2}"), we were able to make two observations. First, we find that Transformer (T5) based model generates asymmetric sentences (Figure \ref{fig:asymmetric}); i.e., the first part of the output, which is related to \textit{target\_1}, is longer than the second par generated from \textit{target\_2}. Second,  the first part of the output covers structured inputs better than the second part. We call this phenomenon \textbf{Asymmetric Generation}. 

Asymmetric Generation can be exploited to improve the recall mentioned earlier. Based on our experiments on ToTTo corpus \citep{parikh2020totto} and WIKITABLET \citep{chen2020wikitablet}, the first part of asymmetric output shows a higher recall than the second part. It is even higher than the output of the model trained by no-repeated-targets. 

We concatenate the first part with a no-repeated-target ("the first part <SEP> no-repeated-target"), then train a new model with lengthened targets (Figure \ref{fig:proedit-3}). This process can be conducted repeatedly. We call this process \textbf{Pro}gressive \textbf{Edit} (ProEdit)\footnote{The name of our model comes from ProGen \citep{tan2020progressive}} because it progressively edits the initial output. Our experimental results on ToTTo corpus demonstrate the benefit of ProEdit in achieving the new state-of-the-art on PARENT \citep{dhingra2019handling} metric.

\begin{table*}[t]
\begin{center}
\scriptsize
\begin{tabular}{|c|ccccc|ccccc|}
\hline
               & \multicolumn{5}{c|}{ToTTo (input=418.43)}                                                                                          & \multicolumn{5}{c|}{WIKITABLET (input=412.34)}                                                     \\
\textbf{T5-large} & \textbf{BLEU$\uparrow$} & \textbf{P$\uparrow$} & \textbf{R$\uparrow$}                     & \textbf{F1}\tablefootnote{we used the official scripts \url{https://github.com/google-research-datasets/ToTTo}}$\uparrow$                    & \textbf{Length (ref=86.4)} & \textbf{BLEU$\uparrow$} & \textbf{P$\uparrow$} & \textbf{R$\uparrow$} & \textbf{F1}\tablefootnote{we used the official scripts \url{https://github.com/mingdachen/WikiTableT}}$\uparrow$ & \textbf{Length (ref=627)} \\ \hline
No-Repeated-Target       & 49.3           & 80.21       & 50.80                           & 58.53                           & 80.39            & 20.05
          & 56.44       & 23.78       & 32.61        & 391.30            \\ \hline
Asymmetric Generation-First  & 44.2           & 77.91       &  52.56 & 59.23 & 94.18            & \textbf{23.7}         & 52.22       & 25.21       & \textbf{33.05}        & 532.33            \\
Asymmetric Generation-Second & 29.2           & 66.62       & 34.88                           & 41.72                           & 92.11            & 14.47          & 45.39       & 17.78       & 24.50        &457.72            \\ \hline
ProEdit-1-First  & 43.6           & 78.13       &  54.00 & \textbf{60.43} & 97.62            & 23.47         & 49.82       & \textbf{25.77}       & 32.93        & 617.76            \\
ProEdit-1-Second & 35.6           & 74.96       & 38.34                           & 46.33                           & 76.19            & 14.47          & 45.39       & 17.78       & 24.50        &457.72            \\ \hline
ProEdit-2-First  & 42.1           & 77.61       &  \textbf{54.24} & 60.39 & 101.174            & -        & -       & -       & -        & -            \\
ProEdit-2-Second & 29.2           & 66.62       & 34.88                           & 41.72                           & 92.11            & -          & -       & -       & -        & -           \\ \hline
\end{tabular}

\end{center}
\caption{Results on ToTTo and WIKITABLET validation set evaluated by BLEU and PARENT. Asymmetric Generation is trained with repeated target sentences that are divided by a special token. Asymmetric Generation-First is the output before <SEP> token. Asymmetric Generation-Second is the output after <SEP> token. ProEdit-1 is trained with Asymmetric Generation-First concatenated to a no-repeated-target. ProEdit-1-First is the output before <SEP> token. ProEdit-2 is trained in the same process using the output from ProEdit-1. P, R, F1 are PARENT precision, recall, and F1 score, respectively}
\label{tab:asymmetric}
\end{table*}

\section{Related Work}

\textbf{PARENT Metric.} The PARENT metric is introduced to evaluate generated texts from structured inputs automatically.

Its precision for n-gram, denoted by \(E_{p}^{n}\), is given by:\\ 

\noindent\(E_{p}^{n}=\)

\begin{equation} \label{eq1}
\frac{\sum_{g\in G_{n}^{i}}^{}[Pr(g\in R_{n}^{i})+Pr(g\notin R_{n}^{i})w(g)]\#G_{n}^{i}(g)}{\sum_{{g\in G_{n}^{i}}}\#G_{n}^{i}(g)}
\end{equation}
 
where  \(R_{n}^{i},G_{n}^{i} \) denote the collection of n-grams of order n in \(R_{}^{i}\) and \(G_{}^{i} \), which are i-th targets and generated texts, respectively. \(Pr(g\in R_{n}^{i})\) is given by \(\frac{\#G_{n}^{i},R_{n}^{i}(g)}{\#G_{n}^{i}(g)}\), where \(\#G_{n}^{i}(g)\) is the count of n-gram \(g\) in \( G_{n}^{i}\)
and \(\#G_{n}^{i},R_{n}^{i}(g)\) denotes the minimum of its counts in \( G_{n}^{i}\) and \(R_{n}^{i}\). Entailment probability, denoted as \(w(g)\), is the most important and further introduced to check whether the presence of an n-gram \(g\) in a generated text is "correct" given structured inputs. Two models have been introduced to calculate of \(w(g)\): the Word overlap model, and the Co-occurrence model. In the most cases, the Word overlap model is used, so we also used it too. 

The recall of PARENT is computed against both the target \(E_{r}(R_{}^{i})\) and the table  \(E_{r}(T_{}^{i})\). These are combined for geometric average:
 \begin{equation}
     E_{r}=E_{r}(R_{}^{i})_{}^{1-\lambda}E_{r}(T_{}^{i})_{}^{\lambda}
 \end{equation}

\(E_{r}(R_{}^{i})\) compute the recall of generated sentences against target sentences. \(E_{r}(T_{}^{i})\) is computed as follows:
\begin{equation}\label{tablerecall}
    E_{r}(T_{}^{i}) = \frac{1}{K}\sum_{k=1}^{K}\frac{1}{|\bar{r_{k}}|}LCS(\bar{r_{k}},G_{}^{i})
\end{equation}
 
K denotes the number of records in table and \(\bar{r_{k}}\) is the number of token in the value string of a record. \(LCS(x,y)\) denotes the length of the longest common subsequence between x and y. The hyperparameter \(\lambda\) can be obtained heuristically using Eq.\ref{tablerecall}
. The key idea is that if the recall of the target against the table is high, it already covers most of the information of structured inputs, so we can assign it a big weight (\(1-\lambda)\). We use this method.

\section{Progressive Edit of Text}

We add the first part of asymmetric generation to an existing target and trained a new model by using this dataset. This process is reiterated so that after the output of the previous model is added to an existing target,  a new model is trained by this dataset. As a result the output sentence goes through progressive editing. (Figure \ref{fig:proedit-3}). ProEdit was repeated until the PARENT metric increased.

\section{Experiment}
\subsection{Dataset and Implementation details}
We used a pre-trained T5-large model with 737.66M parameters. To test Asymmetric Generation, we used two datasets in D2T: the entire set of ToTTo and the part of WIKITABLET. The ToTTo Dataset is composed of Wikipedia tables paired with human-written descriptions. WIKITABLET, which combines Wikipedia table data with its corresponding Wikipedia sections, is similar to ToTTO, but has a long target text. These datasets were selected since their lengthened targets do not exceed the maximum length of T5 decoder. 

On ToTTo, the training set was made up of 120k examples, and the validation set had 7.7k examples. In the case of WIKITABLET, we used 100k and 2.7k samples for training and validation, respectively. As the length of the encoder input was 512. structured inputs exceeding the encoder input were cut to 512. We set the batch size to 2 and learning rate of 5e-5. In the generation phase, we used beam search of size 5, and early stopping with no repeat ngram size 7. Experiments were conducted with two A100 GPUs.

\subsection{Results}
When there were two repeated target sentences (that were divided by a <SEP> token), the generated output was not the same (Table \ref{tab:asymmetric}); this is the Asymmetric Generation phenomenon. Although target sentences are simply repeated, the resulting sentences differed in length and performances on metrics. On both datasets, the first generated part was longer than the second and performances on metrics were higher for the former than for the after. In addition, the results of the first part on the PARENT metric were even better than when using no-repeated-targets on both datasets. 

We put the first part of the output with repeated-targets for the first place of the iteration; this is ProEdit. ProEdit is repeated until the overall F1 score goes up. The result is organized in Table \ref{tab:asymmetric}. The overall F1 score reached the highest with ProEdit-1-First for ToTTo. For WIKITABLET, ProEdit-1-First F1 score was higher than the no-repeated-target, but lower than the Asymmetric Generation-First. In both datasets, the recall values of the first part increased steadily as ProEdit was repeated.

Our method was evaluated on the test set of ToTTo (Table \ref{tab:testset}). Using ProEdit-1-First, we achieved the state-of-the-art on PARENT score for the test set.






\begin{table}
\begin{center}
\scriptsize
\begin{tabular}{llll}
\cline{1-3}
                                                 \textbf{Model}    & \textbf{BLEU$\uparrow$} & \textbf{PARENT$\uparrow$}                &  \\ \cline{1-3}
Pointer-Generator \citep{see2017get} & 41.6           & 51.6                            &  \\
T5-based \citep{kale2020text}        & 49.5           & 58.4                            &  \\
PlanGen \citep{su2021plan}           & 49.2           & 58.7                            &  \\ \cline{1-3}
Ours (ProEdit-1-First)                                            & 48.6           & \textbf{59.18} &  \\ \cline{1-3}
\end{tabular}

\end{center}
\caption{Evaluation results of BLEU and PARENT on the Test set of ToTTo. All the results are cited from the official leaderboar (\url{https://github.com/google-research-datasets/ToTTo}) of ToTTo.}
\label{tab:testset}
\end{table}

\section{Analysis}

\textbf{Recall and Length.} In general, a longer sentence increases the recall value. The first part of Asymmetric Generation and ProEdit have longer sentences than the reference.  For fair comparison, similar output lengths are generated using beam search and minimum length settings (Table \ref{tab:length}). For beam search, the longest sentence is selected. However, selecting the longest sentence in the beam search rather reduced the recall value. In minimum length settings, the probability of the EOS token is zero until the output reaches a certain length. Setting the minimum length improves both length and recall values, but dramatically reduces precision. which led to a steady decrease in the overall F1 score. Our proposed method can improve the overall F1 score since the precision decreases slightly.



\noindent\textbf{Asymmetric Generation}. We conducted repeated target sentences experiment using ToTTo on another model: GPT-2\citep{radford2019language}. Asymmetric Generation also occurs (Table \ref{tab:gpt2}), so it is not a phenomenon that occurs only in the T5 model.

\begin{table}

\begin{center}
\scriptsize

\begin{tabular}{cccccc}
\hline
         \textbf{T5-large} & \textbf{BLEU$\uparrow$} & \textbf{P$\uparrow$} & \textbf{R$\uparrow$} & \textbf{F1$\uparrow$}    & \textbf{\begin{tabular}[c]{@{}c@{}}Length\\ (ref=86.4)\end{tabular}} \\ \hline
No-Repeated-\\Target  & \textbf{49.3}          & 80.21      & 50.80      & \textbf{58.53}          & 80.39                                                                \\ \hline
Beam size 5 & 45.8          & 77.78      & 49.57      & 56.86          & 88.24                                                                \\
Beam size 11 & 44.3          & 77.02      & 49.24      & 56.48           & 90.65                                                                \\\hline
min length 20 & 45.9          & 77.78      & 51.32      & 58.17 & 89.27  \\
min length 25 & 40 & 74.48 & 52.13 & 57.61 & 101.98\\ 
min length 30 & 34.6          & 71.22      & \textbf{52.68}      & 56.74 & 116.79  \\

\end{tabular}

\end{center}
\caption{Results on ToTTo validation set evaluated by BLEU and PARENT with official scripts.}

\label{tab:length}
\end{table}





\begin{table}

\begin{center}
\scriptsize

\begin{tabular}{cccccc}
\hline
         \textbf{GPT2} & \textbf{BLEU$\uparrow$} & \textbf{P$\uparrow$} & \textbf{R$\uparrow$} & \textbf{F1$\uparrow$}    & \textbf{\begin{tabular}[c]{@{}c@{}}Length\\ (ref=86.4)\end{tabular}} \\ \hline
No-Repeated-\\Target  & 42.8          & 79.13      & 43.49      & 51.74          & 76.26                                                                \\ \hline
Asymmetric\\ Generation-First & 33.2          & 74.97      & \textbf{47.57}      & \textbf{53.89}          & 110.43                                                                \\\hdashline
Asymmetric\\ Generation-Second & 23.4          & 64.45      & 30.82      & 37.16           & 103.29                                                                \\\hline

\end{tabular}

\end{center}
\caption{Results on ToTTo validation set evaluated by BLEU and PARENT with official scripts using GPT2.}

\label{tab:gpt2}
\end{table}

\section{Conclusion}
In this paper, we proposed Progressive Edit (ProEdit) process for D2T generation. It utilizes Asymmetric Generation to improve recall. We obtained the new state-of-the-art result on ToTTo. 

\bibliography{anthology,custom}
\bibliographystyle{acl_natbib}

\appendix

\label{sec:appendix}
\onecolumn

\section{Examples of generated results}
\begin{table}[hb!]
\resizebox{\textwidth}{!}{%
\begin{tabular}{|cl|}

\hline
\multicolumn{2}{|c|}{\textbf{Input Table}}                                                                                                                                                                                                                                                                                                                                                       \\ \hline
\multicolumn{2}{|l|}{\begin{tabular}[c]{@{}l@{}}\textbf{Page\_Title}{[}Hudson Line (Metro-North){]} \textbf{Section\_Title}{[}Stations{]}  \textbf{Zone}{[}Harlem–125th Street \\ Handicapped/disabled access{]} \textbf{Zone}{[}Harlem / New Haven Lines diverge{]} \textbf{Station Miles (km) from GCT}  \\ \textbf{Date opened Date closed Manhattan / Bronx border Zone}{[}Yankees–East 153rd Street Handicapped/disabled access{]}\end{tabular}} \\ \hline
\multicolumn{2}{|c|}{Target Sentence}                                                                                                                                                                                                                                                                                                                                                            \\ \hline
\multicolumn{2}{|l|}{\begin{tabular}[c]{@{}l@{}}Once past 125th Street and the Harlem, the Hudson Line departs from the Harlem and New Haven Lines, passing first \\ Yankees–East 153rd Street.\end{tabular}}                                                                                                                                                                                    \\ \hline
\multicolumn{1}{|c|}{\textbf{No-Repeated-Target}}                                                  & \begin{tabular}[c]{@{}l@{}}The Harlem / New Haven Lines diverge at 125th Street and Yankees–East \\ 153rd Street.\end{tabular}                                                                                                                                                              \\ \hline
\multicolumn{1}{|c|}{\textbf{Asymmetric Generation-First}}                                         & \begin{tabular}[c]{@{}l@{}}The Harlem–125th Street and Yankees–East 153rd Street stations are\\  in the Hudson Line (Metro-North)\end{tabular}                                                                                                                                              \\ \hline
\multicolumn{1}{|c|}{\textbf{ProEdit-1-First}}                                                     & \begin{tabular}[c]{@{}l@{}}The Harlem–125th Street and the New Haven Lines diverge from the Harlem\\  / New Haven Lines to the Yankees–East 153rd Street Handicapped/disabled \\ access stations of the Hudson Line (Metro-North) line.\end{tabular}                                        \\ \hline
\end{tabular}%
}
\end{table}

\begin{table}[hb!]
\resizebox{\textwidth}{!}{%
\begin{tabular}{|cl|}
\hline
\multicolumn{2}{|c|}{\textbf{Input Table}}                                                                                                                                                                                                         \\ \hline
\multicolumn{2}{|l|}{\begin{tabular}[c]{@{}l@{}} \textbf{Page\_Title}{[}Sunda Kingdom{]} \textbf{Section\_Title}{[}List of monarchs{]} \textbf{Period}{[}723 – 732{]}\\  \textbf{King's name}{[}Sanjaya/Harisdarma/ Rakeyan Jamri{]} \textbf{Ruler}{[}Sunda, Galuh, and Mataram{]}\end{tabular}} \\ \hline
\multicolumn{2}{|c|}{Target Sentence}                                                                                                                                                                                                              \\ \hline
\multicolumn{2}{|l|}{In 723, Jamri was the King of Sunda.}                                                                                                                                                                                         \\ \hline
\multicolumn{1}{|c|}{\textbf{No-Repeated-Target}}            & Sanjaya (r. 723–732) was the ruler of the Sunda Kingdom.                                                                                                                            \\ \hline
\multicolumn{1}{|c|}{\textbf{Asymmetric Generation-First}}   & In 723, Sanjaya, Harisdarma and Rakeyan Jamri were the rulers of the Sunda Kingdom.                                                                                                 \\ \hline
\multicolumn{1}{|c|}{\textbf{ProEdit-1-First}}               & \begin{tabular}[c]{@{}l@{}}Sanjaya/Harisdarma/Rakeyan Jamri was the king of the Sunda Kingdom from 723 to 732, \\ ruling from the Sunda, Galuh, and Mataram dynasty.\end{tabular}   \\ \hline
\end{tabular}%
}
\end{table}

\begin{table}[hb!]
\resizebox{\textwidth}{!}{%
\begin{tabular}{|cl|}
\hline
\multicolumn{2}{|c|}{\textbf{Input Table}}                                                                                                                                                                           \\ \hline
\multicolumn{2}{|l|}{\begin{tabular}[c]{@{}l@{}}\textbf{Page\_Title}{[}Herculaneum, Missouri{]} \textbf{Section\_Title}{[}Demographics{]} \textbf{Historical population}{[}2010{]}\\  \textbf{Census Historical population}{[}3,468{]} \textbf{Pop}\end{tabular}} \\ \hline
\multicolumn{2}{|c|}{Target Sentence}                                                                                                                                                                                \\ \hline
\multicolumn{2}{|l|}{As of the census of 2010, there were 3,468 people residing in Herculaneum, Missouri.}                                                                                                           \\ \hline
\multicolumn{1}{|c|}{\textbf{No-Repeated-Target}}                                            & The population of Herculaneum was 3,468 at the 2010 census.                                                           \\ \hline
\multicolumn{1}{|c|}{\textbf{Asymmetric Generation-First}}                                   & As of the census of 2010, there were 3,468 people residing in the Herculaneum.                                        \\ \hline
\multicolumn{1}{|c|}{\textbf{ProEdit-1-First}}                                               & As of the census of 2010, there were 3,468 people residing in Herculaneum, Missouri.                                  \\ \hline
\end{tabular}%
}
\caption{Examples of generated result on the ToTTo dataset}
\end{table}

\newpage
\section{Rules for post-processing}
When lengthened targets were given, but divided by a <SEP> token, the output of the model sometimes did not produce a <SEP> token or generated several. We divided the first part and the second part into the following rules.

\(\circ\) If <SEP> does not occur:

\begin{table}[h]
\begin{center}
\begin{tabular}{|l|}
\hline
generated sentence \\ \hline
\end{tabular}
\end{center}
\end{table}

the first part = generated sentence

the second part = generated sentence

\(\circ\) If [SEP] occurs once:

\begin{table}[h]
\begin{center}
\begin{tabular}{|l|}
\hline
\begin{tabular}[c]{@{}l@{}}generated sentence\_1 {[}SEP{]} \\ generated sentence\_2\end{tabular} \\ \hline
\end{tabular}

\end{center}
\end{table}

the first part = generated sentence\_1

the second part = generated sentence\_2

\(\circ\) If [SEP] occurs serveral:
\begin{table}[h]
\begin{center}
\begin{tabular}{|l|}
\hline
\begin{tabular}[c]{@{}l@{}}generated sentence\_1 {[}SEP{]}\\ generated sentence\_2 {[}SEP{]}\\ generated sentence\_3 {[}SEP{]}\\ ...\end{tabular} \\ \hline
\end{tabular}

\end{center}
\end{table}

the first part = generated sentence\_1

the second part = generated sentence\_1

\end{document}